\documentclass[10pt,journal,compsoc]{IEEEtran}
\usepackage{amsmath,amsfonts}
\usepackage{algorithmic}
\usepackage{algorithm}
\usepackage{array}
\usepackage{textcomp}
\usepackage{stfloats}
\usepackage{url}
\usepackage{verbatim}
\usepackage{graphicx}
\usepackage{cite}
\usepackage{amsmath}
\usepackage{amsfonts}  
\usepackage{amsthm,amssymb}
\usepackage{bm} 
\usepackage{booktabs} 
\usepackage{multirow}
\usepackage{xcolor} 

\usepackage{dutchcal}
\usepackage{makecell}
\usepackage{threeparttable}
\usepackage[T1]{fontenc}
\usepackage{booktabs}
\usepackage{blkarray}

\usepackage{mathrsfs}
\usepackage{subfigure}
\usepackage[hang]{footmisc}
\usepackage[font=small,labelfont=bf]{caption}
\usepackage[graphicx]{realboxes}
\usepackage{epstopdf}

\begin{document}

\title{Joint Learning of Unsupervised Multi-view Feature and Instance Co-selection with Cross-view Imputation}

\author{Yuxin Cai, Yanyong Huang, Jinyuan Chang, Dongjie Wang, Tianrui Li,~\IEEEmembership{Senior Member,~IEEE}, and Xiaoyi Jiang,~\IEEEmembership{Senior Member,~IEEE}
\thanks{Yuxin~Cai, Yanyong~Huang, and Jinyuan Chang are with the Joint Laboratory of Data Science and Business Intelligence, School of Statistics and Data Science, Southwestern University of Finance and Economics, Chengdu 611130, China (e-mail: 1230202j8005@smail.swufe.edu.cn; huangyy@swufe.edu.cn; changjinyuan@swufe.edu.cn), Yanyong Huang is the corresponding author;}

\thanks{Dongjie Wang is with the Department of Electrical Engineering and Computer Science, University of Kansas, Lawrence, KS 66045, USA (e-mail: wangdongjie@ku.edu);}

\thanks{Tianrui Li is with the School of Computing and Artificial Intelligence, Southwest Jiaotong University, Chengdu 611756, China (e-mail: trli@swjtu.edu.cn);}

\thanks{Xiaoyi Jiang is with the Faculty of Mathematics and Computer Science, University of Münster, Münster 48149, Germany (e-mail: xjiang@uni-muenster.de).}}

\markboth{Journal of \LaTeX\ Class Files,~Vol.~14, No.~8, August~2021}%
{Shell \MakeLowercase{\textit{et al.}}: A Sample Article Using IEEEtran.cls for IEEE Journals}

\IEEEpubid{0000--0000/00\$00.00~\copyright~2021 IEEE}

\maketitle

\begin{abstract}
Feature and instance co-selection, which aims to reduce both feature dimensionality and sample size by identifying the most informative features and instances, has attracted considerable attention in recent years. However, when dealing with unlabeled incomplete multi-view data, where some samples are missing in certain views, existing methods typically first impute the missing data and then concatenate all views into a single dataset for subsequent co-selection. Such a strategy treats co-selection and missing data imputation as two independent processes, overlooking potential interactions between them. The inter-sample relationships gleaned from co-selection can aid imputation, which in turn enhances co-selection performance. Additionally, simply merging multi-view data fails to capture the complementary information among views, ultimately limiting co-selection effectiveness. To address these issues, we propose a novel co-selection method, termed Joint learning of Unsupervised multI-view feature and instance Co-selection with cross-viEw imputation (JUICE). JUICE first reconstructs incomplete multi-view data using available observations, bringing missing data recovery and feature and instance co-selection together in a unified framework. Then, JUICE leverages cross-view neighborhood information to learn inter-sample relationships and further refine the imputation of missing values during reconstruction. This enables the selection of more representative features and instances. Extensive experiments demonstrate that JUICE outperforms state-of-the-art methods.
\end{abstract}

\begin{IEEEkeywords}
Feature and instance co-selection, Unlabeled incomplete multi-view data, Data reconstruction.
\end{IEEEkeywords}

\section{Introduction}
\begin{figure*}[!htbp]
	\centering 
	\includegraphics[width=1\textwidth]{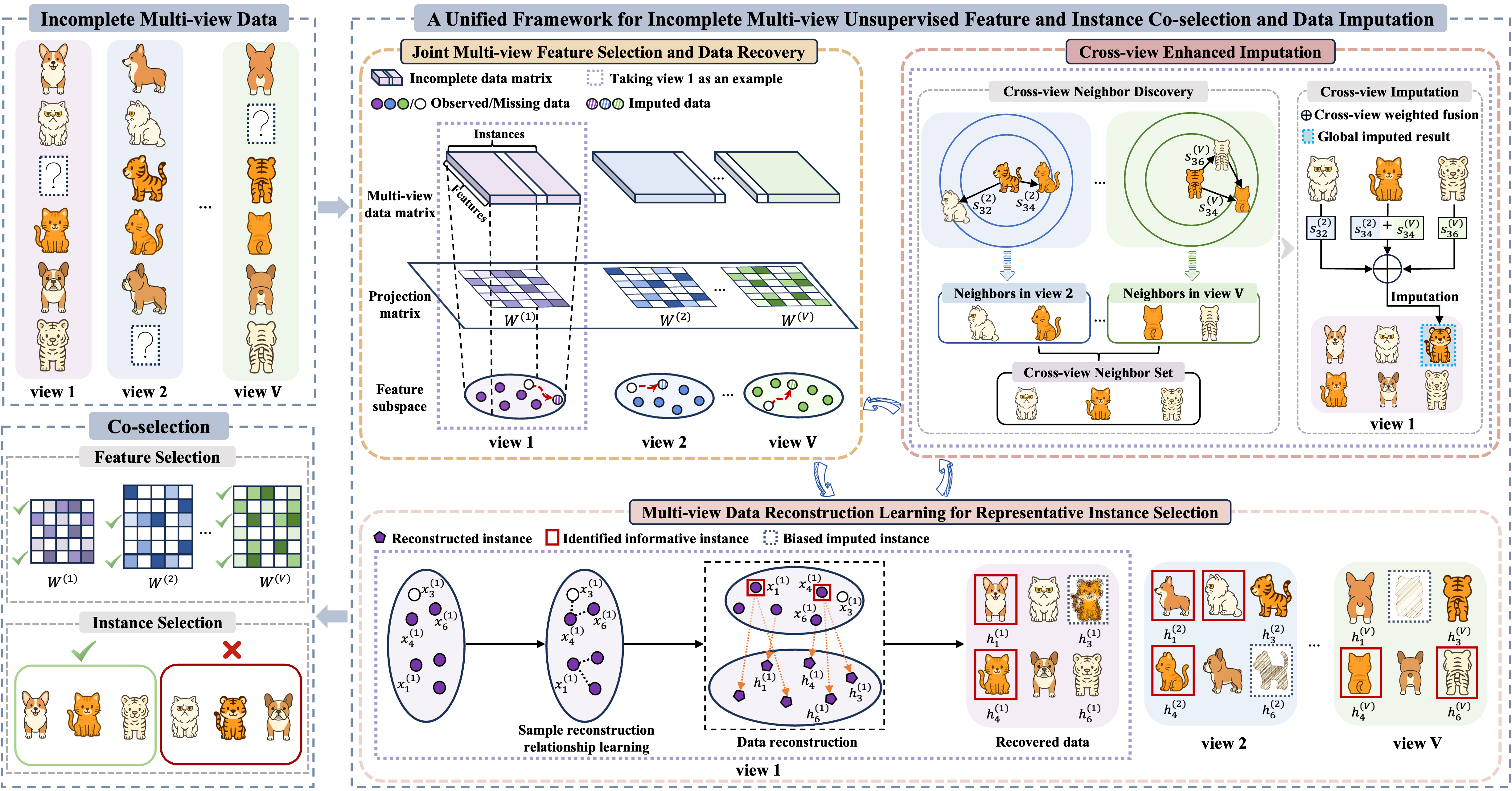}
	\caption{The framework of the proposed JUICE.}
	\label{overall framework}
\end{figure*}

The rapid advancement of information technology has made multi-view data increasingly common, in which heterogeneous features from different perspectives are used to describe the same sample~\cite{CDSL_EAAI2024, ASFL_PR2025, JMVFG_TETCI2023}. Obtaining large amounts of labeled multi-view data in real-world applications is often difficult or even impossible. Unlabeled multi-view data typically contains numerous redundant features and instances, which will result in the curse of dimensionality and degraded model performance~\cite{AMVFS_ICDM2023, BSFRS_TFS2023}. Feature and instance co-selection, which jointly reduces feature dimensionality and sample size by identifying the most informative features and representative instances, has emerged as an effective approach for the preprocessing of unlabeled multi-view data.

Existing feature and instance co-selection methods for unlabeled multi-view data can generally be divided into two categories. The first category combines multi-view unsupervised feature selection (MUFS) and instance selection methods to sequentially select features and instances from multi-view data. Typical MUFS methods include C$^2$IMUFS~\cite{C2IMUFS_TKDE2023} and UKMFS~\cite{UKMFS_AAAI2025}, while representative instance selection methods include CIS~\cite{CIS_IS2022} and NIS~\cite{NIS_NCA2023}. As a result, these approaches treat feature and instance selection as separate processes and fail to exploit the potential interactions between their respective spaces, which may limit performance. In contrast to the first category, which uses a combination-based approach, the second category concatenates features from all views into a single set and subsequently employs traditional single-view-based co-selection methods, such as UFI~\cite{UFI_TIP2012}, DFIS~\cite{DFIS_Access2020}, and sCOs2~\cite{sCOs2_TKDE2022}. sCOs2 incorporates pairwise similarity preservation and an $\ell_{2,1-2}$-norm regularization term to simultaneously enable the selection of discriminative features and instances.

Although these methods can somewhat improve co-selection performance, most of them assume that every sample is fully observed in all views. This assumption may not always hold true in the real-world scenario. For example, in environmental monitoring, some stations have only temperature and humidity data, and lack volatile organic compound measurements due to the high cost of installing specialized analytical instruments~\cite{VOC}. When dealing with incomplete multi-view data, existing co-selection methods typically employ a two-step strategy: first, imputing missing data with fixed values (such as the mean), and then performing co-selection. However, treating data imputation and co-selection as two independent processes fails to account for their potential synergy. The inter-sample relationships gleaned from co-selection are beneficial for imputation, which in turn further boosts co-selection performance.
In addition, these single-view-based co-selection methods just combine multi-view data by simply concatenating them, treat each view separately, and overlook the relationships between different views, which results in suboptimal co-selection performance.

To address the above problems, we propose a novel unsupervised feature and instance co-selection method for incomplete multi-view data, termed Joint learning of Unsupervised multI-view feature and instance Co-selection with cross-viEw imputation (JUICE). Specifically, we first reconstruct the multi-view data in a projected low-dimensional space using available observations, which facilitates the simultaneous identification of informative features and representative instances. Meanwhile, missing data are recovered by exploiting reconstruction relationships among samples and are integrated with multi-view co-selection within a unified framework, enabling mutual reinforcement between these two processes. Then, we use cross-view neighborhood information to perform a weighted fusion of reconstructed data for each view, wherein the weights are determined by neighborhood relationships identified in different views. The resulting fused data is subsequently utilized to refine the imputation of missing values and guide the reconstruction of sample relationships within each view, thereby improving the effectiveness of co-selection. Fig.~\ref{overall framework} illustrates the overall framework of the proposed JUICE. The main contributions of this paper are summarized as follows:
\begin{itemize}
	\item To the best of our knowledge, this is the first work to propose and study unsupervised feature and instance co-selection for incomplete multi-view data, which advances co-selection to a more realistic scenario. 
	\item The proposed JUICE utilizes both cross-view and intra-view information to effectively integrate multi-view unsupervised feature and instance co-selection and missing data recovery into a joint learning framework, wherein these modules collaboratively enhance the co-selection performance.
	\item An effective iterative algorithm is designed to optimize the proposed JUICE, and comprehensive experimental results convincingly demonstrate the superiority of JUICE over state-of-the-art methods.
\end{itemize}

The remainder of this paper is organized as follows. Section 2 provides a brief review of related work on feature and instance co-selection. Section 3 details the proposed JUICE method. In Section 4, we describe the optimization algorithm for JUICE, while Section 5 discusses its complexity and convergence. Section 6 presents promising comparative results on real-world multi-view datasets. Section 7 concludes the paper.

\section{Related Work}
Existing unsupervised multi-view feature and instance co-selection methods are typically classified into two main categories. The first category utilizes a combined strategy that sequentially applies MUFS and instance selection. MUFS methods commonly leverage inter-view relationships to identify a compact and informative subset of features from multi-view unlabeled data, including representative approaches such as UKMFS~\cite{UKMFS_AAAI2025} and SCMvFS~\cite{SCMvFS_ESWA2024}. UKMFS projects the original data into a latent kernel subspace and adaptively learns a consensus graph using hash-based binary labels to guide feature selection. SCMvFS constructs a similarity graph for each view and applies spectral analysis to these graphs, enabling the derivation of a consensus spectral embedding that facilitates the selection of informative features. Given that some instances may be missing in certain views, several incomplete MUFS methods have been developed. UIMUFSLR~\cite{UIMUFSLR_TII2025} integrates multi-view graph fusion with low-redundancy learning into a unified framework, enabling simultaneous missing data recovery and feature selection for unbalanced incomplete data. TERUIMUFS~\cite{TERUIMUFS_IF2025} employs tensor low-rank representation combined with sample diversity learning to recover missing instances, while leveraging self-representation to identify important features. C$^2$IMUFS~\cite{C2IMUFS_TKDE2023} utilizes cross-view complementary information to learn a complete similarity graph for each view and derives a consensus indicator matrix from the view-specific graphs to guide feature selection. Moreover, instance selection serves as the dual problem to feature selection, with a primary focus on identifying representative samples from the original dataset to mitigate noise and enhance the performance of downstream tasks. Representative approaches include CIS~\cite{CIS_IS2022} and NIS~\cite{NIS_NCA2023}. CIS presents a cluster-oriented approach that utilizes k-means clustering to select instances from both cluster centers and boundaries. NIS implements a data scaling technique to partition the dataset into hyperrectangles, retaining a single sample from each region for instance selection. 
The first category of methods combines MUFS with instance selection to enable joint selection of features and instances. However, these approaches treat feature selection and instance selection as separate processes, neglecting the interaction between them, which ultimately limits their performance.

The second category of methods adopts a co-selection framework that simultaneously identifies informative features and representative instances, rather than relying on the combined strategy utilized by the first category. Representative methods in this category include UFI~\cite{UFI_TIP2012}, DFIS~\cite{DFIS_Access2020}, and sCOs2~\cite{sCOs2_TKDE2022}. UFI adopts a greedy optimization strategy to jointly select informative features and representative instances through the minimization of the parameter covariance matrix trace. DFIS employs a data reconstruction technique to simultaneously select representative features and instances, effectively preserving the intrinsic manifold structure of the data. sCOs2 integrates pairwise similarity preservation with an $\ell_{2,1-2}$-norm regularization term within a unified framework to facilitate simultaneous feature and instance selection. Although the above-mentioned methods have demonstrated promising performance in feature and instance co-selection, they still encounter notable limitations when deployed on incomplete multi-view datasets. Specifically, these methods generally assume that each sample is observed in all views. When handling incomplete data, they typically employ a sequential two-step strategy: missing samples are first imputed, followed by feature and instance co-selection. However, this sequential strategy fails to account for the potential interactions between the imputation and co-selection processes. Moreover, these approaches are primarily designed for single-view datasets.  When extended to multi-view datasets, they typically concatenate  all views into a single dataset for co-selection, which results in treating each view independently and ignoring the underlying relationships between views. This can negatively affect the performance of feature and instance co-selection.

\section{The Proposed Method} \label{Method}
\subsection{Notations and Problem Definition}
Throughout this paper, both matrices and vectors are denoted by bold uppercase letters, while scalars are written in regular font. For any matrix  $\bm{X} \in \mathbb{R}^{d \times n}$, $\bm{X}_{i\cdot}$, $\bm{X}_{\cdot j}$, and $X_{ij}$ denotes the  $i$-th row, $j$-th column and  $(i,j)$-th entry of $\bm{X}$, respectively. The transpose of $\bm{X}$ is denoted by $\bm{X}^{T}$, and its trace by $Tr( \bm{X})$. The $\ell_{2}$-norm of $\bm{X}_{i\cdot}$ is defined as $\| \bm{X}_{i\cdot} \|_{2} = \sqrt{\sum_{j=1}^{n}X_{i j}^{2}}$. The Frobenius norm of  $\bm{X}$ is given by $\| \bm{X}\|_{F}=\sqrt{\sum_{i=1}^{d}\sum_{j=1}^{n}X_{i j}^{2}}$, and the  $\ell_{2,1}$-norm of $\bm{X}$ is defined as $\| \bm{X}\|_{2,1}=\sum_{i=1}^{d}\sqrt{\sum_{j=1}^{n}X_{i j}^{2}}$. $\bm{I}$ represents the identity matrix, and $\bm{1}$ denotes a column vector whose entries are all equal to one.

To formalize the problem, we consider an incomplete multi-view dataset $\mathcal{X} = \{\bm{X}^{(v)}  \in \mathbb{R}^{d_{v}\times n} \}_{v=1}^V$ with $V$ views, where $\bm{X}^{(v)}$ is the data matrix for view $v$, containing $d_{v}$ features and  $n$ instances. In order to describe the missing data setting, in which some instances are absent from certain views, we introduce an indicator matrix $\bm{G}^{(v)} \in \mathbb{R}^{n \times n_{v}} (v=1,\cdots, V)$ for each view, where  $G_{ij}^{(v)}=1$ if the $j$-th observed instance corresponds to $\bm{X}_{\cdot i}^{(v)}$, and $G_{ij}^{(v)}=0$ otherwise. The observed data matrix $\bm{O}^{(v)} $ of the $v$-th view can then be expressed  as  $\bm{O}^{(v)} = \bm{X}^{(v)} \bm{G}^{(v)}\in \mathbb{R}^{d_{v}\times n_{v} }$, where $n_v$ denotes the number of available instances in the $v$-th view. This study aims to simultaneously select the $h$ most informative features and the $m$ most representative instances from the incomplete multi-view dataset $\mathcal{X}$.

\subsection{Joint Learning of Co-selection and Data Recovery}
Although many joint feature and instance selection methods have been developed for single-view data~\cite{DFIS_Access2020, sCOs2_TKDE2022}, these methods typically address incomplete multi-view data through a two-step strategy. Specifically, missing values are first imputed with fixed values, such as the mean, and then the data are concatenated into a single view for subsequent selection. This strategy treats co-selection and data imputation as independent processes, overlooking potential interactions between them. 
Furthermore, simply concatenating multi-view data does not take inter-view information into account, which may ultimately result in suboptimal joint selection performance.  

To address these issues, we propose a joint learning framework that seamlessly integrates multi-view feature and instance co-selection with missing data recovery. We first reconstruct the data in a projected low-dimensional space using the available observations. During the reconstruction process, we simultaneously select informative features and instances. This can be formulated as follows:
\begin{equation} \label{NSR1}
	\begin{aligned}
		&
		\begin{aligned}
			\min_{\bm{W}^{(v)},\bm{Q}^{(v)}} &\!\!\sum_{v=1}^{V}[ \|\bm{W}^{(v)T} (\bm{X}^{(v)} \!\!-\! \bm{O}^{(v)} \bm{Q}^{(v)T})\|_{F}^{2} \!+\!\! \lambda \| \bm{W}^{(v)} \|_{2,1}  \\
			& + \alpha \| \bm{Q}^{(v)} \|_{F}^{2} ]
		\end{aligned} \\
		&~~~~s.t.\bm{W}^{(v)T} \bm{W}^{(v)} = \bm{I}, \bm{Q}^{(v)} \bm{1} =  \bm{1},  \bm{Q}^{(v)} \geq 0,Q^{(v)}_{i^{'}i} = 0, \\
		& ~~\qquad v=1,2,...,V, 
	\end{aligned}
\end{equation}	
where $\bm{W}^{(v)}  \in  \mathbb{R}^{d_{v} \times c}$ denotes the feature selection matrix, $\bm{Q}^{(v)} \in  \mathbb{R}^{n \times n_v}$ denotes the instance selection matrix,  and $\alpha$  and $\lambda$ are regularization parameters. Here, $\bm{O}^{(v)}=\bm{X}^{(v)} \bm{G}^{(v)}$ indicates the observed data in the $v$-th view. In Eq.~(\ref{NSR1}), the first term projects the original high-dimensional data onto a low-dimensional space and reconstructs it using the available data, thereby reducing the influence of irrelevant or noisy features and facilitating the identification of representative instances. The $\ell_{2,1}$-norm is imposed on $\bm{W}^{(v)}$ to encourage row sparsity and enable the identification of informative features. In addition, the $i$-th column of $\bm{Q}^{(v)}$ represents the weights assigned by the $i$-th instance to reconstruct the other instances. By imposing the Frobenius norm on $\bm{Q}^{(v)}$, this regularization encourages Eq.~(\ref{NSR1}) to select similar samples for reconstruction.  If the elements in $\bm{Q}_{\cdot i}^{(v)}$ are large, it indicates that the $i$-th instance is highly similar to the others, suggesting that this sample is more important. We also impose the constraint $Q^{(v)}_{i^{'}i}=0~(i^{'} \in \{ j \mid G^{(v)}_{ji} = 1 \})$ to ensure that each sample cannot be reconstructed by itself. Therefore, we can use the $\ell_{2}$-norm of each column in $\bm{Q}^{(v)}$ to select representative instances.

Due to the presence of missing data in $\bm{X}^{(v)}$, the first term in Eq.~(\ref{NSR1}) cannot be used to compute the reconstruction error, which limits its ability to effectively guide the joint selection of features and instances. To address this, we introduce a reconstructed data matrix $\bm{H}^{(v)} \in \mathbb{R}^{d_{v} \times n}$ that allows for the reconstruction of all instances and the imputation of missing values, while ensuring consistency between the reconstructed and observed data. This process can be formulated as follows:
\begin{equation} \label{NSR2}
	\begin{aligned}
		\min_{\Psi} 
		&\!\!\sum_{v=1}^{V} \!\! \frac{1}{\gamma^{(v)}}  [  \| \bm{W}^{(v)T} (\bm{H}^{(v)} \!-\! \bm{O}^{(v)} \bm{Q}^{(v)T} ) \|_{F}^{2} \!+\! \lambda \| \bm{W}^{(v)} \|_{2,1}\\
		&  + \alpha  \| \bm{Q}^{(v)} \|_{F}^{2} + \beta \| \bm{H}^{(v)} \bm{G}^{(v)} - \bm{O}^{(v)} \|_{F}^{2} ],\\
		s.t. & ~ \bm{W}^{(v)T} \bm{W}^{(v)} = \bm{I}, \bm{\gamma} \bm{1} = 1,  \bm{\gamma} \geq 0, \bm{Q}^{(v)} \bm{1} = \bm{1},\\
		& ~\bm{Q}^{(v)} \geq 0, Q^{(v)}_{i^{'}i} = 0, v=1,2,...,V\\
	\end{aligned}
\end{equation} 
where $\Psi \!=\! \{\bm{H}^{(v)}, \bm{Q}^{(v)}, \bm{W}^{(v)}, \gamma^{(v)} \}_{v=1}^{V}$, $\gamma^{(v)}$ is the weight factor of the $v$-th view, and  $\bm{\gamma} = [\gamma^{(1)}, \gamma^{(2)}, \dots, \gamma^{(V)}]$ denotes the weight factor vector for different views. In Eq.~(\ref{NSR2}), the first term imputes missing values by reconstructing each sample based on the available observations, while the last term encourages the reconstructed data to closely align with the original data at the observed entries. Moreover, we introduce $\bm{\gamma}$ to adaptively compute the weight for each view, thereby facilitating the integration of information from different views. Eq.~(\ref{NSR2}) integrates multi-view feature and instance co-selection with missing data recovery into a unified learning framework, enabling them to enhance each other. According to Eq.~(\ref{NSR2}),  the top $h$ features can be selected by ranking the $\ell_2$-norms of the rows of $\{\bm{W}^{(v)}\}_{v=1}^{V}$ in descending order. In addition, we evaluate the importance of the $i$-th instance by calculating a weighted sum of the $\ell_{2}$-norm of $\bm{Q}^{(v)}_{\cdot i}$ using the weight factor $\gamma^{(v)}$, as given by $\sum_{v=1}^{V} \frac{1}{\gamma^{(v)}} \| (\bm{Q}^{(v)} \bm{G}^{(v) T})_{\cdot i} \|_2^{2}$. The instances are then ranked in descending order according to their importance scores, and the top $m$ representative instances are selected.

\subsection{Cross-view Information-enhanced Reconstruction}
Multi-view data characterizes the same instance from multiple perspectives, with each view offering complementary information. Previous studies have shown that integrating cross-view information can substantially enhance model performance~\cite{COMVSC_TKDE2020, LMVC_NN2024,CFSMO_KBS2024}. To this end, we utilize cross-view neighborhood information to carry out weighted fusion of the reconstructed data obtained using only intra-view information, and subsequently employ the resulting data to further refine the imputation of missing values. Specifically, for each view, we first construct the similarity graph $\bm{S}^{(v)}$ as described in~\cite{GMC_TKDE2019}. Using these graphs, we identify the $k$-nearest neighbors and determine the similarity between samples for each view. These similarity values then serve as weights in the weighted fusion of the reconstructed data matrix $\bm{H}^{(v)}$ obtained from Eq.~(\ref{NSR2}). The fused results are then used to update the missing values in the data matrix $\bm{X}^{(v)}$. This process can be described as follows:
\begin{equation} \label{NX3}
	\bar{\bm{X}}^{(v)}_{\cdot i}=M_{iv} \bm{X}^{(v)}_{\cdot i} + (1-M_{iv}) \sum_{u=1}^{V} \frac{ \bm{H}^{(v)} \bm{S}^{(u)T}_{i \cdot}}{ \sum_{j=1}^{n}\sum_{u=1}^{V}  \bm{S}^{(u)}_{ij}},
\end{equation}
where $M_{iv}$ is the $(i,v)$-th entry of the index matrix $\bm{M} \in \mathbb{R}^{n \times V}$, with $M_{iv} = 1$ if instance $i$ is observed in view $v$, and $M_{iv} = 0$ otherwise. According to Eq.~(\ref{NX3}), we can obtain the data matrix $ \{\bar{\bm{X}}^{(v)} \in \mathbb{R}^{d_v \times n} \}_{v=1}^{V} $ refined by cross-view information. Moreover, the refined data are further used to guide the reconstruction of relationships among samples within each view, as formulated below:
\begin{equation}  \label{NX4}
	\min_{\bm{Q}^{(v)}} \quad \sum_{i=1}^{n} \sum_{j=1}^{n_v} Q_{ij}^{(v)} \| \bm{\bar{X}}_{\cdot i}^{(v)} - \bm{O}_{\cdot j}^{(v)} \|_{2}^{2},
\end{equation}
where $\bm{\bar{X}}_{\cdot i}^{(v)}$ and $\bm{O}_{\cdot j}^{(v)}$ represent the $i$-th and $j$-th instances of $\bm{\bar{X}}^{(v)} $ and $\bm{O}^{(v)} $, respectively. Eq.~(\ref{NX4}) indicates that the greater the distance between the observed instance $\bm{O}_{\cdot j}^{(v)}$ and the instance $\bm{\bar{X}}_{\cdot i}^{(v)}$ (i.e., the less similar they are), the smaller the weight $Q_{ij}^{(v)}$ assigned to using $ \bm{O}_{\cdot j}^{(v)}$ to reconstruct $\bm{\bar{X}}_{\cdot i}^{(v)}$ will be. By integrating Eqs.~(\ref{NX3}) and (\ref{NX4}), we can leverage cross-view information to regulate each observed sample's contribution during reconstruction, thereby facilitating the selection of more representative instances.

By combining Eqs.~(\ref{NSR2}), (\ref{NX3}), and (\ref{NX4}), the final objective function of the proposed JUICE is formulated as follows:
\begin{equation}  \label{general objFunc}
	\begin{aligned}
		\min_{\Omega} &
		\!\!\sum_{v=1}^{V} \!\! \frac{1}{\gamma^{(v)}}  [  \| \bm{W}^{(v)T} (\bm{H}^{(v)} \!-\! \bm{O}^{(v)} \bm{Q}^{(v)T} ) \|_{F}^{2} \!+\!\! \lambda \| \bm{W}^{(v)} \|_{2,1} \\
		& + \alpha \| \bm{Q}^{(v)} \|_{F}^{2} + \beta \| \bm{H}^{(v)} \bm{G}^{(v)} - \bm{O}^{(v)} \|_{F}^{2} \\
		& + \sum_{i=1}^{n} \sum_{j=1}^{n_v} Q_{ij}^{(v)} \| \bm{\bar{X}}_{\cdot i}^{(v)} \!-\! \bm{O}_{\cdot j}^{(v)} \|_{2}^{2}  ]\\
		s.t. & \bm{W}^{(v)T} \bm{W}^{(v)} = \bm{I}, \bm{\gamma} \bm{1} = 1,  \bm{\gamma} \geq 0, \bm{Q}^{(v)} \bm{1} = \bm{1},\\
		& \bm{Q}^{(v)} \geq 0, Q^{(v)}_{i^{'}i}  = 0, v=1,2,...,V\\
	\end{aligned}
\end{equation}
where $\Omega=\{\bm{H}^{(v)}, \bm{Q}^{(v)}, \bm{W}^{(v)}, \bm{\bar{X}}^{(v)},\gamma^{(v)} \}_{v=1}^{V}$. On the one hand, the proposed JUICE seamlessly integrates multi-view feature and instance co-selection with missing data recovery into a joint learning framework. By selecting more informative features and representative instances for data reconstruction, JUICE not only facilitates data recovery and captures relationships among samples more effectively, but also enhances the performance of multi-view co-selection in turn. On the other hand, it leverages cross-view information to refine imputed data and adjust the contribution of observable instances during reconstruction, thereby making more effective use of both inter-view and intra-view information for data reconstruction.

\section{Optimization and Algorithm}
Given that the objective function in Eq.~(\ref{general objFunc}) is non-convex with respect to all variables, we propose an alternating iterative algorithm that sequentially optimizes each variable while keeping the others fixed.

\noindent \textbf{Updating $\bm{H}^{(v)}$:}
With other variables fixed, $\bm{H}^{(v)}$ can be updated by solving the following optimization problem:
\begin{equation} \label{H objFunc}
	\begin{aligned}
		\min_{\bm{H}^{(v)}} 
		& \| \bm{W}^{(v)T} (\! \bm{H}^{(v)} \!\!-\! \bm{O}^{(v)} \bm{Q}^{(v)T}\! )  \|_{F}^{2} \!+\!\! \beta  \| \bm{H}^{(v)} \bm{G}^{(v)} \!\!-\! \bm{O}^{(v)} \! \|_{F}^{2} \\
		&  + \sum_{i=1}^{n} \sum_{j=1}^{n_v} Q_{ij}^{(v)} \| \bm{\bar{X}}_{\cdot i}^{(v)} - \bm{O}_{\cdot j}^{(v)} \|_{2}^{2}.
	\end{aligned}
\end{equation}

By taking the derivative of Eq. (\ref{H objFunc}) with respect to  $\bm{H}^{(v)}$ and setting it to zero, we obtain:
\begin{equation} \label{Hv update}
	\bm{A}^{(v)} \bm{H}^{(v)} \!+\! \bm{H}^{(v)} \bm{B}^{(v)} \!=\! \bm{C}^{(v)},
\end{equation}
where $\bm{A}^{(v)} \!=\! \bm{W}^{(v)} \bm{W}^{(v)T}$, 
$\bm{B}^{(v)}\!=\! \bm{S}^{(v)T} \bm{U}^{(v)} \bm{U}^{(v)} \bm{S}^{(v)} + \beta \bm{G}^{(v)} \bm{G}^{(v)T}$, and the $i$-th diagonal entry of the diagonal matrix $\bm{U}^{(v)}$ is $\frac{1 \!-\! M_{iv}}{\sum_{v=1}^{V} \! \bm{S}^{(v)}_{i \cdot} \bm{1}}$. The matrix $\bm{C}^{(v)}$ is defined as $ \bm{W}^{(v)} \bm{W}^{(v)T}  \bm{O}^{(v)} \bm{Q}^{(v)T} \!+\!  \beta  \bm{O}^{(v)} \bm{G}^{(v)T} \!-\!  \bm{Z}^{(v)}  \bm{U}^{(v)}  \bm{S}^{(v)}$, where the $i$-th column of $ \bm{Z}^{(v)}$ is given by $ M_{iv} \bm{X}^{(v)}_{\cdot i} - \sum_{j=1}^{n_v} Q_{ij}^{(v)} \bm{O}^{(v)}_{\cdot j}$. This equation is a generalized Sylvester equation. Following~\cite{CVFL_AAAI2014, MFSICC_TSMCS2023}, it can be rewritten as $(\bm{I}_{n} \otimes \bm{A}^{(v)} \!+\! \bm{B}^{(v)T} \otimes \bm{I}_{d_v}) vec(\bm{H}^{(v)}) \!=\! vec(\bm{C}^{(v)})$, where $vec( \cdot )$ denotes the vectorization operator, $\otimes$ is the Kronecker product, $\bm{I}_n$ and $\bm{I}_{d_v}$ represent the identity matrices of dimensions $n$ and $d_v$, respectively. Since the Kronecker product increases rapidly with matrix dimensions, leading to substantial computational and storage overheads, the Bi-Conjugate Gradient (BiCG) algorithm~\cite{BCG_convergence} is employed to compute $\bm{H}^{(v)}$.

\noindent\textbf{Updating $\bm{Q}^{(v)}$:}
When the other variables are fixed, the optimization subproblem for $\bm{Q}^{(v)}$ can be solved row by row, as each row is independent. It can be reformulated as follows:
\begin{equation} \label{Qi objFunc}
	\begin{aligned}
		\min_{\bm{Q}^{(v)}_{i \cdot}} 
		&~ \| \bm{E}^{(v)}_{\cdot i} - \bm{F}^{(v)} \bm{Q}_{i \cdot }^{(v)T} \|_{2}^{2} 
		+ \| \bm{Q}^{(v)}_{i \cdot} + \frac{1}{2\alpha} \bm{N}^{(v)}_{i \cdot} \|_{2}^{2}  \\
		s.t. 
		&\bm{Q}^{(v)} \bm{1} = \bm{1}, \bm{Q}^{(v)} \geq 0,  Q^{(v)}_{i^{'}i}  = 0,
	\end{aligned}
\end{equation}
where $\bm{E}^{(v)} \!=\! \bm{W}^{(v)T} \bm{H}^{(v)}$, $\bm{F}^{(v)} =\bm{W}^{(v)T} \bm{O}^{(v)}$ and the $(i,j)$-th entry of $\bm{N}^{(v)}$ is given by $\| \bm{\bar{X}}_{\cdot i}^{(v)} \!-\! \bm{O}_{\cdot j}^{(v)} \|_{2}^{2}$. Following~\cite{EMUFS_IJCAI2024, SEMI_IF2024}, we take the derivative of Eq. (\ref{Qi objFunc}) with respect to $\bm{Q}^{(v)}_{i \cdot}$ and obtain an intermediate solution $\hat{\bm{Q}}_{i \cdot}^{(v)}$. Subsequently, the optimal solution of $\bm{Q}_{i \cdot}^{(v)}$ can  be determined by solving:
\begin{equation}  
	\begin{aligned}
		&\min_{\bm{Q}^{(v)}_{i \cdot}} ~~
		\| \bm{Q}^{(v)}_{i \cdot} - \hat{\bm{Q}}^{(v)}_{i \cdot}\|_{2}^{2} \\
		s.t. 
		&\bm{Q}^{(v)} \bm{1} = \bm{1}, \bm{Q}^{(v)} \geq 0,  Q^{(v)}_{i^{'}i}  = 0,
	\end{aligned}
\end{equation}
Following~\cite{GMC_TKDE2019}, we can obtain the following result using the Karush-Kuhn-Tucker (KKT) conditions:
\begin{equation} \label{Qv update}
	Q_{ij}^{(v)} = (R_{ij}^{(v)} - \hat{\psi})_{+},
\end{equation}
where $R_{ij}^{(v)}$ is the $j$-th entry of $\bm{R}_{i \cdot}^{(v)} = \hat{\bm{Q}}_{i \cdot}^{(v)} + \frac{\bm{1}^{T}}{n_v} - \frac{\hat{\bm{Q}}_{i \cdot}^{(v)} \bm{1} \bm{1}^{T}}{n_v}$. Additionally, $\hat{\psi}$ can be computed by applying the Newton method to find the root of the equation $\frac{1}{n_v} \sum_{j=1}^{n_v}{(\hat{\psi}- R_{ij}^{(v)})_{+}} - \hat{\psi}=0$.

\noindent\textbf{Updating $\bm{W}^{(v)}$:}
After fixing other variables, the objective function with respect to $\bm{W}^{(v)}$ is reduced to:
\begin{equation}   \label{Wv update}
	\begin{aligned}
		\min_{\bm{W}^{(v)}}  \quad
		&Tr [\bm{W}^{(v)T} (\bm{P}^{(v)} \bm{P}^{(v)T}+\lambda \bm{D}^{(v)})\bm{W}^{(v)} ] \\
		s.t. \quad 
		& \bm{W}^{(v)T} \bm{W}^{(v)} = \bm{I},
	\end{aligned}
\end{equation}
where $\bm{P}^{(v)}=\bm{H}^{(v)} - \bm{O}^{(v)} \bm{Q}^{(v)T}$. Besides, $\bm{D}^{(v)}$  is a diagonal matrix, and its $i$-th diagonal element is given by  $D^{(v)}_{ii} = \frac{1}{2\|\bm{W}^{(v)}_{i \cdot} \|_{2}+\epsilon}$, where $\epsilon$ is a sufficiently small constant introduced to avoid division by zero. The optimal $\bm{W}^{(v)}$ in Eq.~(\ref{Wv update}) is formed by the $c$ eigenvectors of $(\bm{P}^{(v)} \bm{P}^{(v)T}+\lambda \bm{D}^{(v)})$ corresponding to  the smallest $c$ eigenvalues.

\noindent\textbf{Updating $\bm{\bar{X}}^{(v)}$ and $\gamma^{(v)}$:}
When other varibles are fixed, $\bm{\bar{X}}^{(v)}$ can be updated using Eq. (\ref{NX3}). Additionally, by applying  the Lagrange multiplier method to Eq. (\ref{general objFunc}) w.r.t. $\gamma^{(v)}$, we can obtain the optimal solution  for $\gamma^{(v)}$ as follows:
\begin{equation} \label{gamma_v update}
	\gamma^{(v)} = \frac{\sqrt{\phi^{(v)}}}{\sum_{v=1}^{V} \sqrt{\phi^{(v)}}},
\end{equation}
where $\phi^{(v)} \!=\! \| \bm{W}^{(v)T} (\bm{H}^{(v)} \!-\! \bm{O}^{(v)} \bm{Q}^{(v)T}) \|_{F}^{2} \!+\! \lambda \| \bm{W}^{(v)} \|_{2,1} \!+\! \alpha \| \bm{Q}^{(v)} \|_{F}^{2} \!+\! \beta \| \bm{H}^{(v)} \bm{G}^{(v)} - \bm{O}^{(v)} \|_{F}^{2} + \sum_{i=1}^{n} \sum_{j=1}^{n_v} Q_{ij}^{(v)} \| \bm{\bar{X}}_{\cdot i}^{(v)} \!-\! \bm{O}_{\cdot j}^{(v)} \|_{2}^{2}$.

Algorithm~\ref{UpdateAlgorithm} summarizes the overall optimization procedure of JUICE. In this algorithm, $\bm{\bar{X}}^{(v)}$ is initialized based on observed instances in $\{\bm{X}^{(v)}\}_{v=1}^{V}$. $\bm{H}^{(v)}$ is initialized using the incomplete data $\bm{X}^{(v)}$ with missing values initially set to zeros. $\bm{W}^{(v)}$ is randomly initialized to ensure orthogonality. $\bm{Q}^{(v)}$ and $\gamma^{(v)}$ are initialized according to Eqs. (\ref{Qv update}) and (\ref{gamma_v update}), respectively.
\begin{algorithm}[t]
	\caption{Iterative algorithm of JUICE}
	\label{UpdateAlgorithm}
	\textbf{Input:} Incomplete multi-view dataset $\mathcal{X} = \{\bm{X}^{(v)}\}_{v=1}^V$ , indicator matrices $\{\bm{G}^{(v)}\}_{v=1}^V$ for each view, and parameters $\alpha$, $\beta$, $\lambda$. 
	\begin{algorithmic}[1]
		\STATE \textbf{Initialize:}  $\{\bm{\bar{X}}^{(v)}, \bm{H}^{(v)}, \bm{W}^{(v)}, \bm{Q}^{(v)}, \gamma^{(v)}\}_{v=1}^V$.
		\WHILE{not convergent}
		\STATE Update $\{\bm{H}^{(v)}\}_{v=1}^V$ by solving Eq. (\ref{Hv update}) ;
		\STATE Update $\{\bm{Q}^{(v)}\}_{v=1}^V$ via Eq. (\ref{Qv update});
		\STATE Update $\{\bm{W}^{(v)}\}_{v=1}^V$ by solving Eq. (\ref{Wv update});
		\STATE Update $\{\bar{\bm{X}}^{(v)}\}_{v=1}^V$ via Eq. (\ref{NX3});
		\STATE Update $\{\gamma^{(v)}\}_{v=1}^V$ via Eq. (\ref{gamma_v update});
		\ENDWHILE \\
	\end{algorithmic}
	\textbf{Output:} The scores for features and instances are calculated and sorted in descending order, and then the top $h$ features and $m$ instances are selected.
\end{algorithm}

\section{Time Complexity and Convergence Analysis}
In Algorithm~\ref{UpdateAlgorithm}, five variables are updated alternately. For updating $\bm{H}^{(v)}$, the time complexity is $\mathcal{O}(l(n d_v + z))$, where $l$ is the number of iterations of the BiCG algorithm, and $z$ denotes the number of nonzero elements in $\bm{I}_{n} \otimes \bm{A}^{(v)} + \bm{B}^{(v)T} \otimes \bm{I}_{d_v}$. For updating $\bm{Q}^{(v)}$, it takes $\mathcal{O}(nn_{v})$ complexity. Updating $\bm{W}^{(v)}$ has a computation complexity of $\mathcal{O}(d_{v}^{2}c)$. For updating $\bar{\bm{X}}^{(v)}$, the time complexity is $\mathcal{O}(n k d_{v})$. The update of $\gamma_{v}$ consists solely of element-wise operations, and thus its computational cost can be ignored. Hence, the total computational complexity of Algorithm~\ref{UpdateAlgorithm} is $\mathcal{O}(\sum_{v=1}^{V}( l(n d_{v} + z) + nn_v+d_{v}^{2}c +n k d_{v}))$.

Since the objective function in Eq. (\ref{general objFunc}) is not jointly convex with respect to $\bm{H}^{(v)}$, $\bm{Q}^{(v)}$, $\bm{W}^{(v)}$, $\bar{\bm{X}}^{(v)}$ and $\gamma^{(v)}$, it is decomposed into five sub-objective problems. The convergence of Algorithm~\ref{UpdateAlgorithm} is subsequently established by proving the convergence of each sub-problem. Specifically, the sub-problem (\ref{H objFunc}) with regards to $\bm{H}^{(v)}$ is addressed using the BiCG algorithm, whose convergence has been validated in~\cite{BCG_convergence}. Moreover, the convergence of $\bm{Q}^{(v)}$, $\bar{\bm{X}}^{(v)}$, and $\gamma^{(v)}$  is ensured by their closed-form solutions presented in Eqs. (\ref{Qv update}), (\ref{NX3}), and (\ref{gamma_v update}), respectively. The convergence of $\bm{W}^{(v)}$ can be obtained according to~\cite{ACSLL_TKDD2023}.  Experimental results in Section~\ref{Experiments} further demonstrate the convergence behavior of Algorithm~\ref{UpdateAlgorithm}.

\section{Experiments} \label{Experiments}
\subsection{Experimental Settings}
\subsubsection{Datasets}
We conduct experiments on eight real-world multi-view datasets, including Yale\footnote{http://www.cad.zju.edu.cn/home/dengcai/Data/FaceData.html}, MSRC-V1~\cite{MSRC}, COIL20~\cite{COIL20}, HandWritten~\cite{HW}, BDGP~\cite{BDGP}, CCV~\cite{CCV_ALOI}, USPS~\cite{USPS}, and ALOI~\cite{CCV_ALOI}. Table~\ref{Dataset} provides detailed information about these datasets. Additionally, following~\cite{IMLBDR_TC2019,DCP_TPAMI2023}, we generate incomplete multi-view datasets by removing a certain percentage of samples from each view as missing data. To evaluate the performance of the proposed method under varying degrees of incompleteness, we conduct experiments with missing data ratios ranging from 10\% to 50\% at intervals of 10\%.

\subsubsection{Compared Methods}
To demonstrate the effectiveness of the proposed JUICE, we compare it with several state-of-the-art (SOTA) approaches, including single-view-based methods for joint feature and instance co-selection such as DFIS~\cite{DFIS_Access2020}, UFI~\cite{UFI_TIP2012}, and sCOs2~\cite{sCOs2_TKDE2022}, as well as methods that combine multi-view feature selection and instance selection. A brief summary of the comparison methods is provided below:

$\bullet$ \textbf{DFIS}: It uses a data reconstruction framework to simultaneously select features and instances while preserving the intrinsic manifold structure of the original data.

$\bullet$ \textbf{UFI}:  It performs feature and instance co-selection through a greedy optimization process by minimizing the trace of the parameter covariance matrix.

$\bullet$ \textbf{sCOs2}: This method introduces a pairwise similarity preservation term by imposing an $\ell_{2,1-2}$-norm to enforce sparsity, thereby enabling the simultaneous selection of features and instances.

$\bullet$ \textbf{C2IN}: The combined method selects features using a complementary and consensus learning-based incomplete multi-view unsupervised feature selection method (C$^2$IMUFS~\cite{C2IMUFS_TKDE2023}), and selects instances using a conjectural hyperrectangle-based instance selection method (NIS~\cite{NIS_NCA2023}).

$\bullet$ \textbf{TERN}: This method combines TERUIMUFS~\cite{TERUIMUFS_IF2025} for feature selection, which leverages tensor low-rank representation and sample diversity learning, with NIS~\cite{NIS_NCA2023} for instance selection, as in C2IN.

$\bullet$ \textbf{TIMC}: It combines a tensor decomposition-based incomplete multi-view unsupervised feature selection method (TIME-FS~\cite{TIMEFS_AAAI2025}) with an instance selection method (CIS~\cite{CIS_IS2022}), which identifies representative samples based on cluster centers and boundaries.

$\bullet$ \textbf{UKMC}: The combined method performs feature selection via a multi-view hashing approach (UKMFS~\cite{UKMFS_AAAI2025}), which learns a consensus graph structure and binary representations across all views, and conducts instance selection using CIS~\cite{CIS_IS2022}, consistent with TIMC.

$\bullet$ \textbf{UIMD}: This method combines an unbalanced incomplete multi-view unsupervised feature selection approach (UIMUFSLR~\cite{UIMUFSLR_TII2025}), which incorporates a low-redundancy constraint, with an instance selection method (D-SMRS~\cite{DSMRS_PR2015}), which employs staged representative prototypes and pruning to remove noisy samples.

$\bullet$ \textbf{SCMD}: It integrates a multi-view feature selection approach (SCMvFS~\cite{SCMvFS_ESWA2024}), which jointly considers graph heterogeneity and indicator consistency, with an instance selection method (D-SMRS~\cite{DSMRS_PR2015}), similar to UIMD.

\begin{table}[t]
	\caption{Dataset description}
	\label{Dataset}
	\setlength{\tabcolsep}{2pt}
	\renewcommand{\arraystretch}{1.1}
	\resizebox{\linewidth}{!}{
		\begin{tabular}{lccccc}
			\toprule
			Datasets & Abbr. & Views & Instances & Features & Classes \\
			\midrule
			Yale & Yale & 4 & 165 & 256/256/256/256 & 15 \\
			MSRC-V1 & MSRC & 6 & 210 & 1302/48/512/100/256/210 & 7 \\
			COIL20 & COIL & 3 & 1440 & 30/19/30 & 20 \\
			HandWritten & HW & 4 & 2000 & 216/76/64/240 & 10 \\
			BDGP & BDGP & 3 & 2500 & 1000/500/250 & 5 \\
			CCV & CCV & 3 & 6773 & 20/20/20 & 20 \\
			USPS & USPS & 2 & 9298 & 256/32 & 10 \\
			ALOI & ALOI & 4 &10800  & 77/13/64/125 & 100 \\ 
			\bottomrule
		\end{tabular}
	}
	
\end{table}

\begin{table*}[h]
	\centering
	\setlength{\tabcolsep}{2.5pt}
	\caption{Performance comparison of different methods on eight datasets in terms of ACC and F1.}
	\label{Incom4_Ins2_Fea2 table}
	\resizebox{\textwidth}{!}{
		\begin{tabular}{l|cccccccc|cccccccc}
		\toprule
		\multirow{2}{*}{Method} 
		& \multicolumn{8}{c|}{ACC(\%)} 
		& \multicolumn{8}{c}{F1(\%)} \\
		\cmidrule(lr){2-9} \cmidrule(lr){10-17}
		& Yale & MSRC & COIL & HW & BDGP & CCV & USPS & ALOI 
		& Yale & MSRC & COIL & HW & BDGP & CCV & USPS & ALOI \\
		\midrule
		
		JUICE 
		& \textbf{29.55} & \textbf{68.45} & \textbf{87.67} & \textbf{87.38} & \textbf{50.00} & \textbf{20.04} & \textbf{50.17} & \textbf{36.73} 
		& \textbf{28.16} & \textbf{70.68} & \textbf{87.58} & \textbf{87.53} & \textbf{50.41} & \textbf{17.77} & \textbf{49.99} & \textbf{41.04} \\
		
		DFIS  
		& 13.18$\bullet$ & \underline{53.21}$\bullet$ & 49.77$\bullet$ & 59.46$\bullet$ & 26.52$\bullet$ & 11.38$\bullet$ & 20.43$\bullet$ & 20.13$\bullet$ 
		& 13.48$\bullet$ & \underline{53.92}$\bullet$ & 48.16$\bullet$ & 60.46$\bullet$ & 18.16$\bullet$ & 4.84$\bullet$  & 19.99$\bullet$ & 19.06$\bullet$ \\
		
		UFI   
		& 11.67$\bullet$ & 41.43$\bullet$ & 34.62$\bullet$ & 59.78$\bullet$ & 33.60$\bullet$ & 11.37$\bullet$ & 12.25$\bullet$ & 3.07$\bullet$  
		& 10.42$\bullet$ & 40.16$\bullet$ & 30.54$\bullet$ & 61.36$\bullet$ & 31.34$\bullet$ & 4.50$\bullet$  & 7.10$\bullet$  & 4.75$\bullet$ \\
		
		sCOs2 
		& 12.88$\bullet$ & 50.60$\bullet$ & 21.96$\bullet$ & 54.88$\bullet$ & 32.40$\bullet$ & 15.48$\bullet$ & 11.18$\bullet$ & 12.23$\bullet$ 
		& 10.93$\bullet$ & 50.69$\bullet$ & 16.18$\bullet$ & 54.62$\bullet$ & 27.42$\bullet$ & 10.63$\bullet$ & 6.93$\bullet$  & 13.05$\bullet$ \\
		
		C2IN  
		& 15.91$\bullet$ & 48.33$\bullet$ & 10.07$\bullet$ & 43.39$\bullet$ & \underline{48.30}$\bullet$ & 11.64$\bullet$ & \underline{43.29}$\bullet$ & 12.34$\bullet$ 
		& 12.96$\bullet$ & 48.03$\bullet$ & 8.39$\bullet$ & 38.47$\bullet$  & \underline{48.58}$\bullet$ & 6.40$\bullet$  & \underline{44.50}$\bullet$ & 9.96$\bullet$ \\
		
		TERN  
		& 17.42$\bullet$ & 34.52$\bullet$ & 7.38$\bullet$ & 38.69$\bullet$  & 46.25$\bullet$ & \underline{17.27}$\bullet$ & 16.31$\bullet$ & 7.21$\bullet$  
		& 17.32$\bullet$ & 37.04$\bullet$ & 9.09$\bullet$ & 34.18$\bullet$  & 46.64$\bullet$ & \underline{15.69}$\bullet$ & 15.45$\bullet$ & 7.21$\bullet$ \\
		
		TIMC  
		& 20.80$\bullet$ & 38.11$\bullet$ & 65.35$\bullet$ & \underline{75.22}$\bullet$ & 31.50$\bullet$ & 17.00$\bullet$ & 31.54$\bullet$ & \underline{25.12}$\bullet$
		& 19.05$\bullet$ & 38.38$\bullet$ & 63.39$\bullet$ & \underline{75.13}$\bullet$ & 30.26$\bullet$ & 11.44$\bullet$ & 30.23$\bullet$ & \underline{26.07}$\bullet$ \\
		
		UKMC  
		& \underline{22.24}$\bullet$ & 40.24$\bullet$ & 54.97$\bullet$ & 66.01$\bullet$ & 25.43$\bullet$ & 16.14$\bullet$ & 39.39$\bullet$ & 12.19$\bullet$ 
		& \underline{21.60}$\bullet$ & 40.88$\bullet$ & 62.72$\bullet$ & 66.12$\bullet$ & 20.99$\bullet$ & 9.57$\bullet$  & 38.34$\bullet$ & 13.87$\bullet$ \\
		
		UIMD  
		& 15.15$\bullet$ & 31.55$\bullet$ & \underline{73.35}$\bullet$ & 54.19$\bullet$ & 31.55$\bullet$ & 14.93$\bullet$ & 27.05$\bullet$ & 11.74$\bullet$ 
		& 10.53$\bullet$ & 30.72$\bullet$ & \underline{74.33}$\bullet$ & 61.45$\bullet$ & 30.09$\bullet$ & 11.35$\bullet$ & 23.48$\bullet$ & 14.00$\bullet$ \\
		
		SCMD  
		& 9.09$\bullet$ & 49.88$\bullet$ & 46.39$\bullet$ & 65.09$\bullet$ & 36.06$\bullet$ & 16.80$\bullet$ & 28.46$\bullet$ & 12.47$\bullet$ 
		& 7.13$\bullet$ & 49.40$\bullet$ & 46.78$\bullet$ & 64.56$\bullet$ & 33.95$\bullet$ & 11.47$\bullet$ & 24.44$\bullet$ & 14.14$\bullet$ \\
		
		\bottomrule
\end{tabular}
}
\end{table*}

\begin{figure*}[h]  
	\centering 
	\includegraphics[width=0.96\textwidth]{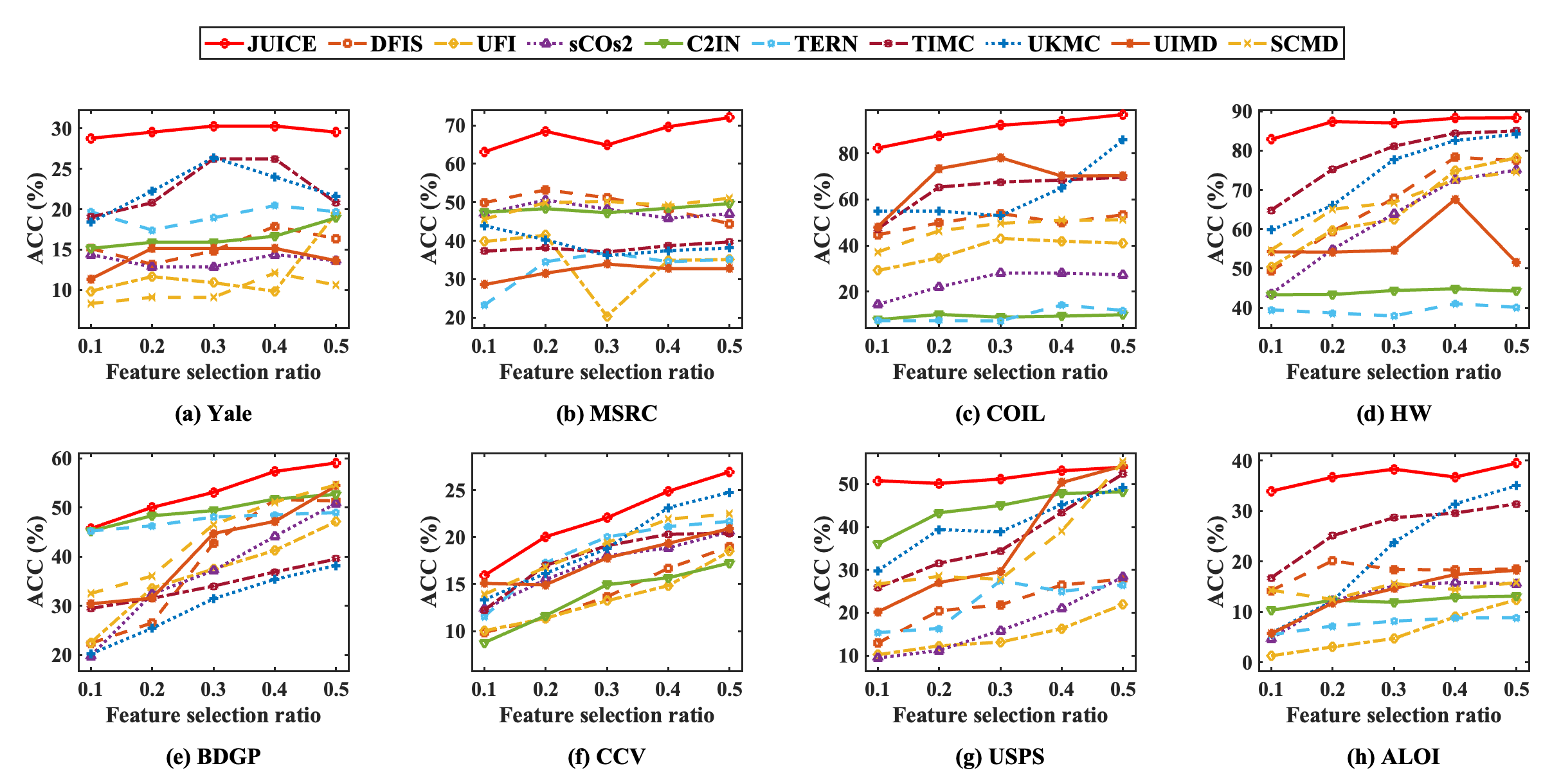}
	\caption{ACC of different methods on eight datasets under different feature selection ratios.}
	\label{FeaVary_ACC_fig}
\end{figure*}

\subsubsection{Comparison Schemes}
Since some comparison methods require complete data and cannot be directly applied to datasets with missing values, we first impute the missing entries using the mean of the corresponding features. Additionally, when applying these single-view-based methods to multi-view data, features from different views are concatenated to form a single-view dataset. For a fair comparison, the parameters of all methods are tuned using grid search, and their best results are reported. In our method, we search for the parameters $\alpha$, $\beta$ and $\lambda$ in $\{10^{-3}, 10^{-2},...,10^{3} \}$. Because of the difficulty in determining the optimal number of features and instances for each dataset~\cite{UFI_TIP2012}, we vary the feature selection ratio and the instance selection ratio from 10\% to 50\% in 10\% increments across all datasets in the experiment. Following previous co-selection methods~\cite{UFI_TIP2012, DFIS_Access2020}, we employ a co-selection approach to identify the most informative features and instances from each dataset. The dataset is subsequently represented by the selected features. 
Following~\cite{OUMDR_TC2022}, we use an SVM as a downstream evaluator to assess the effectiveness of all compared methods. The classifier is trained on the selected instances and their corresponding labels, and is then used to predict the labels of the remaining instances. Classification accuracy (ACC) and F1 score (F1) are used to evaluate the quality of the selected features and instances. Let $y_i$ and $\hat{y}_i$ denote the true and predicted labels of the $i$-th instance, respectively. ACC is defined as:
\begin{equation}
	\text{ACC}=\frac{\sum_{i=1}^{n} \mathbf{1}(\hat{y}_i = y_i)}{n},
\end{equation}
where $\mathbf{1}(\hat{y}_i = y_i)$ is 1 if $\hat{y}_i = y_i$, and 0 otherwise. The F1 score is calculated as:
\begin{equation}
	\text{F1}=\frac{1}{c} \sum_{j=1}^{c}  \frac{2 P_j  R_j}{P_j + R_j},
\end{equation}
where $c$ is the number of classes, $P_j$ and $R_j$ denote the precision and recall for the $j$-th class, respectively. For both metrics, higher values indicate better performance. Each experiment is repeated five times, and the average performance is reported. All experiments were conducted in MATLAB R2022b on a desktop with an Intel Core i9-10900 CPU @ 2.80 GHz and 64 GB RAM.
\begin{figure*}[t]  
	\centering 
	\includegraphics[width=0.95\textwidth]{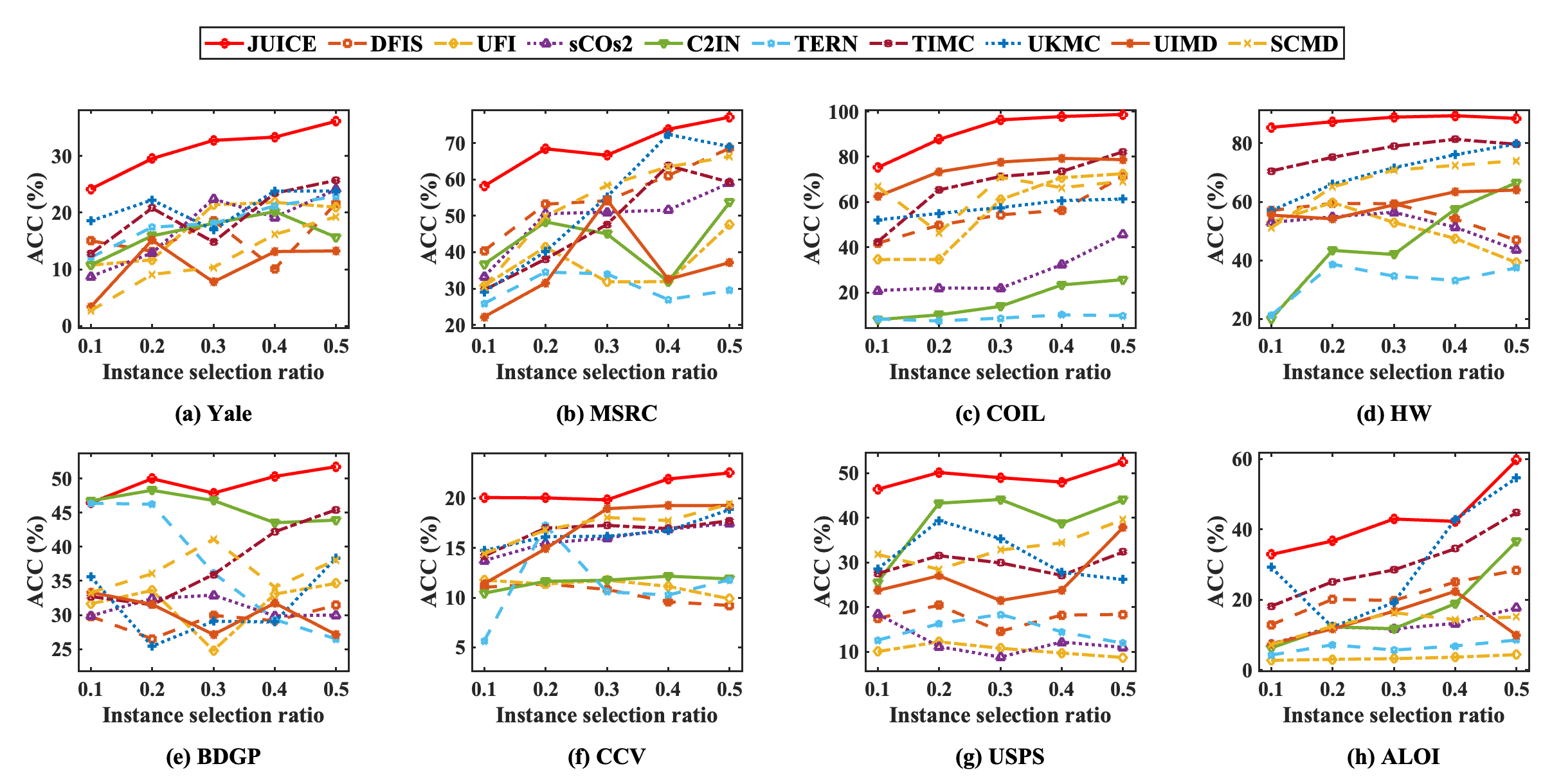}
	\caption{ACC of different methods on eight datasets under different instance selection ratios.}
	\label{InsVary_ACC_fig}
\end{figure*}
\begin{figure*}[!htbp]  
	\centering 
	\includegraphics[width=0.95\textwidth]{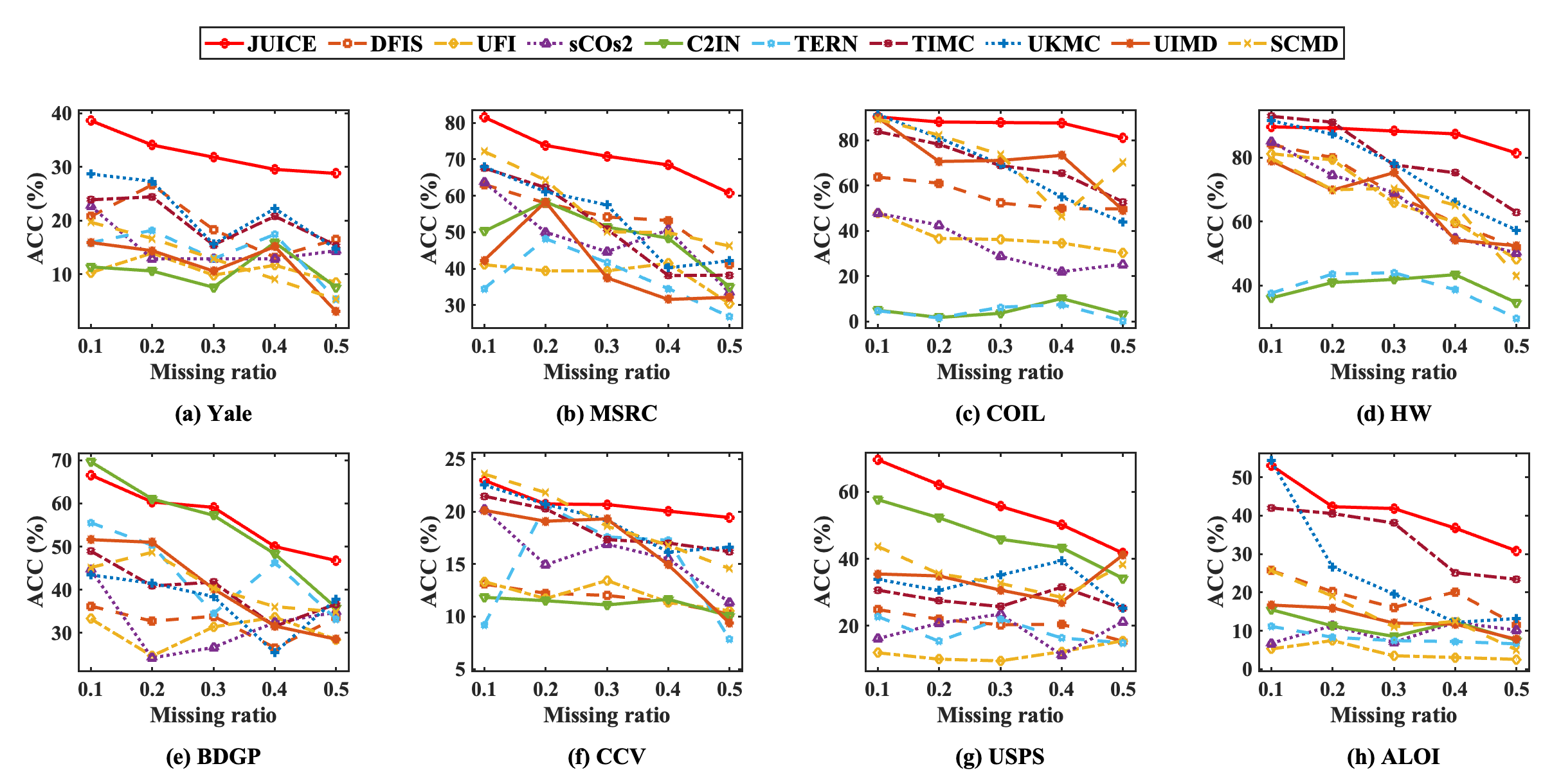}
	\caption{ACC of different methods on eight datasets under different missing ratios.}
	\label{MissingVary_ACC_fig}
\end{figure*}

\subsection{Experimental Results and Analysis}
\subsubsection{Performance Comparison}
Table~\ref{Incom4_Ins2_Fea2 table} presents the performance of the proposed JUICE in comparison with other methods on eight datasets, when 20\% of features and instances are selected and the missing data ratio is set to 40\%. The best and second-best results are shown in bold and underlined, respectively. In addition, we use the Wilcoxon rank-sum test~\cite{ranksumtest} to determine whether the proposed JUICE performs significantly better than other approaches. Values marked with $\bullet$ in Table~\ref{Incom4_Ins2_Fea2 table} show that JUICE achieves statistically significant improvements over other methods at the 0.05 significance level. As shown in the table, the proposed JUICE consistently outperforms the other methods. For the MSRC and COIL datasets, JUICE surpasses the second-best method by more than 13\% in both ACC and F1. As to HW and ALOI datasets, JUICE achieves improvements exceeding 10\% in both metrics. 
On Yale and USPS datasets,  JUICE outperforms the second-best method by more than 6\% in ACC and 5\% in F1. As to BDGP and CCV datasets, JUICE maintains a gain of around 2\% in ACC and F1 relative to the second-best method. In addition, our method outperforms all single-view-based co-selection approaches, achieving an average improvement of more than 10\% in both ACC and F1 across most datasets. Compared to combination-based methods for feature and instance co-selection, our approach also yields an average gain of about 5\% for both metrics. These results demonstrate that jointly leveraging both intra-view and inter-view information enables our method to simultaneously recover missing data and identify informative features and representative instances, thereby outperforming both single-view-based and combination-based approaches.

Furthermore, since it is challenging to determine the optimal number of selected features and instances, we present the performance of all compared methods across different feature and instance selection ratios. Due to space limitations, only the ACC results are shown here, while the F1 results are included in the supplementary material. Figs.~\ref{FeaVary_ACC_fig} and \ref{InsVary_ACC_fig} show the ACC results for varying feature selection ratios and instance selection ratios, respectively, with the missing ratio fixed at 40\%.
As shown in these figures, the proposed JUICE consistently outperforms the other methods across selection ratios ranging from 10\% to 50\% in most cases. Additionally, Fig.~\ref{MissingVary_ACC_fig} presents the ACC results for missing ratios from 10\% to 50\%, with both feature and instance selection ratios fixed at 20\%. JUICE also demonstrates superior performance compared to competing approaches in the majority of cases. The 
F1 scores for varying feature selection ratios, instance selection ratios, and missing ratios can be found in Figs. 1, 2, and 3 in the supplementary material, where JUICE also achieves outstanding performance compared to the other methods in most cases. The superior performance of JUICE can be attributed to its effective integration of multi-view unsupervised feature selection, instance selection, and missing data recovery into a joint learning framework by utilizing cross-view and intra-view information.

\subsubsection{Ablation Study}
\begin{figure}[h]    
	\centering 
	\includegraphics[width=0.46\textwidth]{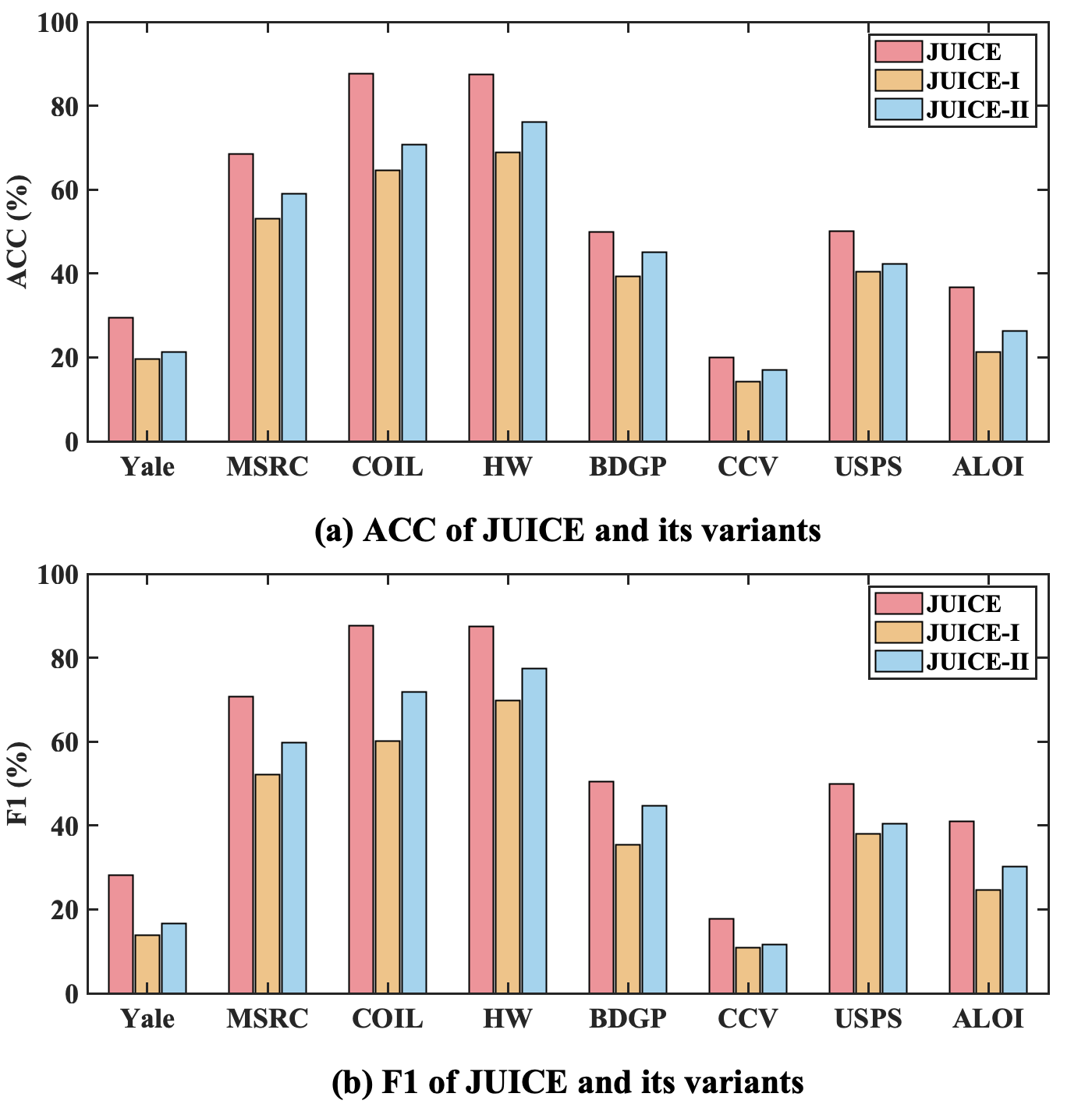}
	\caption{Ablation results of JUICE on eight datasets.}
	\label{ablation results}
\end{figure}

We conduct an ablation study to demonstrate the effectiveness of the proposed module within JUICE. Two variants are developed for comparative analysis: JUICE-$\mathrm{I}$, in which the cross-view information-enhanced module of JUICE is removed; JUICE-$\mathrm{II}$, in which the adaptive missing data recovery module is replaced with mean imputation. Fig. \ref{ablation results} displays the ablation results for JUICE and its two variants on eight datasets in terms of ACC and F1. We can see that the performance of JUICE-$\mathrm{I}$ is significantly lower than that of JUICE, demonstrating the effectiveness of cross-view information in guiding the selection of informative features and representative instances. In addition, JUICE outperforms JUICE-$\mathrm{II}$, confirming the effectiveness of jointly learning co-selection and data recovery.

\subsubsection{Visualization Analysis}
\begin{figure}[t]    
	\centering 
	\includegraphics[width=0.5\textwidth]{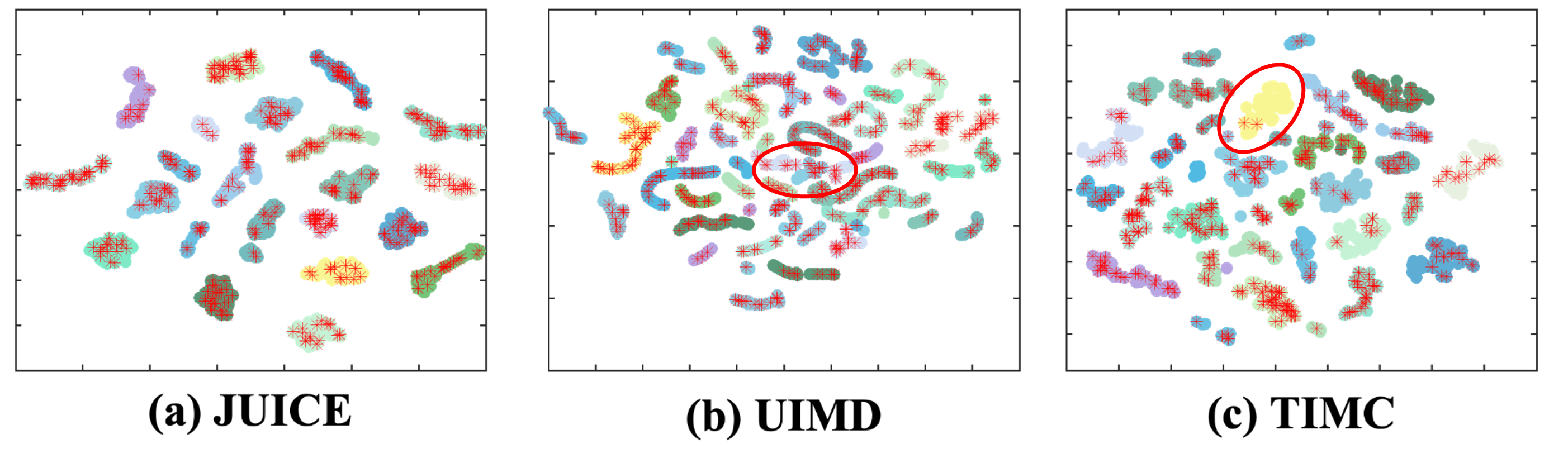}
	\caption{t-SNE visualizations on  COIL dataset.}
	\label{TSNE_COIL_fig}
\end{figure}

To intuitively evaluate the quality of the selected features and instances, we first employ t-SNE~\cite{tSNE} to project all instances based on their selected features into a two-dimensional space, where the selected instances are highlighted with red stars. Fig. \ref{TSNE_COIL_fig} presents the results on the COIL dataset, where 20\% of the features and 30\% of the instances are selected. Fig.~\ref{TSNE_COIL_fig} (a) shows the results of our method, while Fig.~\ref{TSNE_COIL_fig} (b) and Fig.~\ref{TSNE_COIL_fig} (c) correspond to the second- and third-best performing methods, respectively. As can be seen, JUICE yields fairly well-separated clusters with clear boundaries, demonstrating that the selected features are discriminative. Moreover, the identified instances ensure that each cluster is effectively represented, demonstrating its ability to select representative samples from all clusters. In comparison, other methods that use the selected features produce overlapping clusters, and the instances they select fail to represent their corresponding clusters adequately.

\subsubsection{Parameter Sensitivity and Convergence Analysis}
\begin{figure}[!htbp]  
	\centering 
	\includegraphics[width=0.5\textwidth]{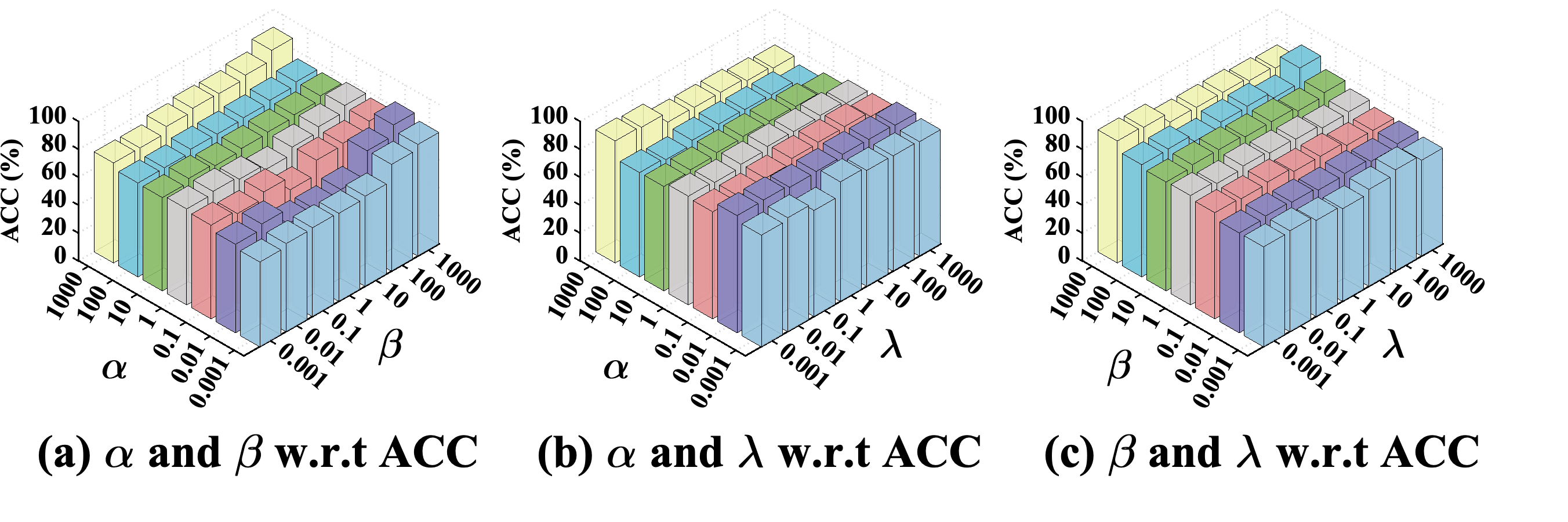}
	\caption{ACC of JUICE with varying parameters $\alpha$, $\beta$ and $\lambda$ on HW dataset.}
	\label{sensitivity_figure}
\end{figure}

\begin{figure*}[t]    
	\centering 
	\includegraphics[width=0.95\textwidth]{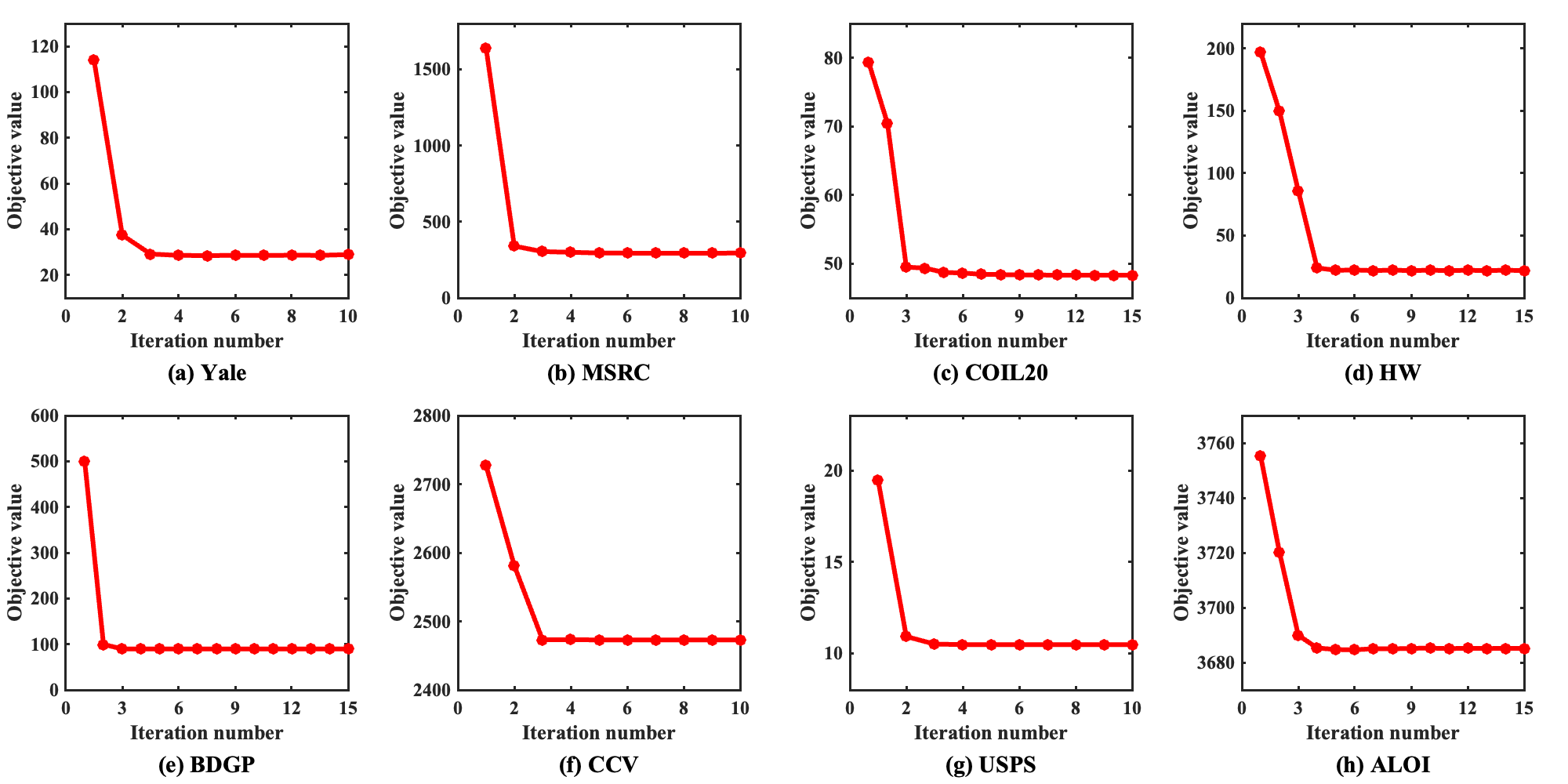}
	\caption{Convergence curves on eight datasets.}
	\label{convergence_figure}
\end{figure*}

The proposed method involves three tuning parameters: $\alpha$, $\beta$, and $\lambda$. We investigate the impact of varying these parameters on the performance of our method. Fig.~\ref{sensitivity_figure} illustrates the sensitivity of our method's ACC to different parameter combinations on HW dataset. As shown in Fig.~\ref{sensitivity_figure}, the ACC exhibits slight fluctuations as parameters $\alpha$, $\beta$, and $\lambda$ vary. In addition, our method generally achieves better performance when $\alpha$ and $\beta$ are relatively large and $\lambda$ is small. Furthermore, Fig.~\ref{convergence_figure} presents the convergence curves of JUICE on eight datasets, showing a sharp decline in the first few iterations and stabilizing after approximately 10 iterations.

\section{Conclusion}
In this paper, we propose a novel unsupervised feature and instance co-selection method, called JUICE, for incomplete multi-view data. Unlike existing approaches that treat co-selection and missing data imputation as separate processes, our method integrates multi-view feature and instance co-selection with missing data recovery into a unified learning framework, enabling mutual enhancement among these components. Furthermore, cross-view neighborhood information is leveraged to collaboratively guide data reconstruction using within-view information, resulting in a more accurate representation of sample relationships and enhanced effectiveness of co-selection. Extensive experimental results demonstrate the effectiveness and superiority of JUICE over SOTA methods.

\bibliographystyle{IEEEtran}
\bibliography{reference}

\vfill

\end{document}